\documentclass{article}

\usepackage{arxiv}

\usepackage[utf8]{inputenc} 
\usepackage[T1]{fontenc}    
\usepackage{hyperref}       
\usepackage{url}            
\usepackage{color}  
\usepackage{booktabs}       
\usepackage{amsfonts}       
\usepackage{nicefrac}       

\usepackage{amsthm}
\usepackage{graphicx}
\usepackage{doi}
\usepackage{physics}
\usepackage{amssymb}
\usepackage{pifont}
\newcommand{\xmark}{\ding{55}}%

\newcommand{\pd}[2]{\frac{\partial #1}{\partial #2}}

\title{On the definition and importance of interpretability in scientific machine learning}

\date{} 					

\author{
Conor Rowan \\
Smead Aerospace Engineering Sciences\\
University of Colorado Boulder\\
3775 Discovery Drive \\
Boulder, CO 80309 \\
USA \\
\texttt{conor.rowan@colorado.edu} \\
\And
Alireza Doostan \\
Smead Aerospace Engineering Sciences\\
University of Colorado Boulder\\
3775 Discovery Drive \\
Boulder, CO 80309 \\
USA \\
\texttt{alireza.doostan@colorado.edu}  
}

\renewcommand{\headeright}{}
\renewcommand{\undertitle}{}
\renewcommand{\shorttitle}{}

\hypersetup{
pdftitle={A template for the arxiv style},
pdfsubject={q-bio.NC, q-bio.QM},
pdfauthor={David S.~Hippocampus, Elias D.~Striatum},
pdfkeywords={First keyword, Second keyword, More},
}

\begin{document}
\maketitle

\begin{abstract}
In the context of scientific machine learning (SciML), the "black box" nature of models involving neural networks makes researchers uneasy. Though neural networks trained on large data sets have been successfully used to describe and predict many physical phenomena, there is a sense that, unlike traditional scientific models---where relationships come packaged in the form of simple mathematical expressions---the findings of the neural network cannot be integrated into the body of scientific knowledge. Critics of machine learning's inability to produce human-understandable relationships have converged on the concept of "interpretability" as its point of departure from more traditional forms of science. As the growing interest in interpretability has shown, researchers in the physical sciences seek not just predictive models, but also to uncover the fundamental principles that govern a system of interest. In hopes of ushering in a future where machine learning (ML) models participate in basic scientific discovery, it is now commonplace to view interpretability as a primary goal. However, clarity around a definition of interpretability and the precise role that it plays in science is lacking in the literature. This is understandable, as concepts such as interpretability and scientific discovery touch on history and philosophy as much as science and ML. In this work, we argue that researchers in equation discovery and symbolic regression tend to conflate the concept of sparsity with interpretability. In search of a more rigorous conception of interpretability, we review key papers on interpretable ML from outside the scientific community and argue that, though the definitions and methods they propose can inform questions of interpretability for SciML, they are inadequate for this new purpose. Noting these deficiencies and drawing from the history and philosophy of science, we propose an operational definition of interpretability that is appropriate for research in the physical sciences. Our notion of interpretability emphasizes understanding of the mechanism over mathematical sparsity, used as a means to \textit{regularize} an inverse/regression problem. Innocuous though it may seem, this emphasis on mechanism shows that, in some cases, sparsity is not an asset. It also questions the possibility of interpretable scientific discovery when prior knowledge is lacking. We believe a precise and philosophically informed definition of interpretability in SciML will help focus research efforts toward the most significant obstacles to realizing a data-driven scientific future.
\end{abstract}



\section{Introduction}

\paragraph{} The triumph of machine learning (ML) models over rule-based systems is a historic event in computing. Until the 1990s, expert systems, which relied entirely on hard-coded rules, were thought to be the frontier of machine intelligence. With initial research beginning in the 1960s, the program "DENDRAL" is widely acknowledged to be one of the first expert systems deployed on a real-world problem. The goal of DENDRAL was to manually encode domain knowledge of organic chemistry to help chemists identify novel compounds from mass spectrometer data \cite{lindsay_dendral_1993}. Another early expert system was "MYCIN," which used hundreds of rules to diagnose the cause of bacterial infections through a series of yes-no questions posed to a physician \cite{shortliffe_mycin_1977}. Though these rule-based methods enjoyed some successes, hope in their viability as a genuine form of machine intelligence quickly declined as the century wore on. This was largely due to the requirement of existing domain knowledge, issues with scalability to more complex problems, and ongoing failures to generalize. In 1997, expert systems suffered a resounding blow when IBM's chess program "Deep Blue" defeated grandmaster Garry Kasparov in a six-game match. Instead of a complex network of hard-coded rules, the domain knowledge that Deep Blue used was an "evaluation function" which scored different board positions \cite{korf_does_1997}. Though the program relied heavily on brute-force search approaches, it still represented a decisive step away from rule-based approaches in computing. 

\paragraph{} In the almost 30 years since Kasparov's defeat, the scientific community has repeatedly seen that ML models often surpass expert systems. Unlike expert systems, ML models do not require manual encoding of domain knowledge in the form of explicit rules. Instead, they find patterns in data to solve problems with comparatively little need for expert intervention. Modern ML began with algorithms such as the support vector machine \cite{vapnik_support_1996} and decision trees \cite{breiman_classification_1984}. By relying only on labeled data and generic mathematical rules for training the model, these methods helped remedy the issues with domain knowledge requirements, scalability, and generalization that expert systems suffered from. In 2012, the field of deep learning was christened by the success of a model called AlexNet, which dramatically outperformed its competitors in an image classification competition \cite{alom_history_2018}. AlexNet was based on a deep convolutional neural network, and convincingly demonstrated the power of these large, black-box models. From this time on, neural networks have become the go-to tool in ML. The all-purpose mathematical framework for function approximation that neural networks provide has encouraged a transition toward ML and big data, and a move away from reliance on explicit rules. Purely data-driven ML models have achieved incredible results in situations where explicit rules would be impossible to enumerate. This includes computer vision, but also applications such as autonomous vehicles \cite{ks_pdf_nodate}, anomaly detection \cite{trilles_anomaly_2024}, speech recognition \cite{ahlawat_automatic_2025}, and drug discovery \cite{yang_pdf_nodate}. Most recently, generative models such as DALL-E and ChatGPT have shown that even art and writing---tasks thought to have closer ties to genuine creativity and human intelligence---can be handled effectively by data-driven ML models. Seeing the promise of these methods elsewhere, scientists have begun to wonder if there are similar opportunities for revolution in their fields.

\paragraph{} In the physical sciences, the systems under study are often governed by ordinary or partial differential equations. In many situations, however, these equations are not precisely known or may be computationally expensive to solve. If it is possible to avoid enumerating rules and teasing out their consequences for tasks such as image classification, might this be possible for physical systems as well? When data relating system inputs (boundary conditions, initial conditions, source terms) to system outputs (temperature, displacement, velocity, etc.) is abundant, modern ML techniques have been deployed to circumvent numerical solutions to the governing differential equations. Initiated by methods like the Deep Operator Network (DeepOnet) \cite{lu_deeponet_2021} and the Fourier Neural Operator (FNO) \cite{li_fourier_2021}, the most popular approach has been "operator learning," where particular neural network architectures are used to represent the dependence of the solution on input fields. This neural network representation of the forward operator is trained on labeled data to map inputs to the output solution field. We view operator learning of this sort as the scientific machine learning (SciML) analogue to the data-driven, neural network-based regression models from the ML community. Other operator learning frameworks have been introduced in the past few years, such as the Neural Green's Operator \cite{melchers_neural_2025, tang_neural_2022} and the Laplace Neural Operator \cite{cao_lno_2023}. Once these models are trained, they can simulate the forward solution to the underlying differential equation at a fraction of the cost, and typically with good accuracy when system inputs are in the interpolation regime of the training data. However, unlike researchers in fields such as computer vision, autonomous vehicles, and natural language processing, scientists are accustomed to systems that can be described with compact mathematical expressions, typically encoding a small set of physical laws. In computer vision, a neural network that classifies images is celebrated based on its accuracy, whereas a scientist may feel predictive accuracy is not the only desideratum of a trained operator network. \textit{A scientist studying a particular physical system demands not just to predict, but also to understand.} The goal of a scientist is to be able to answer the question of: \textit{Why}? Why does the Earth revolve around the sun? Why does a compressively loaded column potentially become unstable? Why does a continuum with material properties fluctuating on small scales admit an effective macroscopic description? These are all questions that a data-driven ML model does not necessarily answer, even when its predictions are extremely accurate. 

\paragraph{} The fields of equation discovery and symbolic regression stand poised to address the critique that operator learning cannot discover a system's underlying mechanisms. With the now-famous equation discovery method titled "sparse identification of nonlinear dynamics" (SINDy), governing differential equations can be learned from data \cite{brunton_discovering_2016}. This represents a departure from directly learning the forward operator of the system, as the goal is not just an instrument for prediction, but rather the true underlying model of the system. SINDy has been extensively studied in the context of many different systems and has spawned a number of variants \cite{shea_sindy-bvp_2021, fries_lasdi_2022, champion_data-driven_2019, delahunt_toolkit_2021}. By enforcing the missing physics to be a sparse combination of terms from a user-defined library, SINDy claims to recover equations which are "interpretable" \cite{wolf_interpretable_2024}. This is taken to be a benefit which transcends the purely mathematical regularization effects of enforcing sparsity \cite{donoho_compressed_2006}. In fact, in this literature, interpretability is championed as the dividing line between historic discoveries such as Kepler's three laws of planetary motion and a learned operator with good generalization properties \cite{brunton_discovering_2016, champion_data-driven_2019}. According to many authors, the importance of interpretability is seen in its demarcating genuine scientific discoveries, but its definition is not precisely specified. For completeness, we note that there are also equation discovery techniques that do not claim to be interpretable. For example, "deep hidden physics" \cite{raissi_deep_2018} and "neural ordinary differential equations" \cite{chen_neural_2019} both represent the system dynamics as a deep neural network. These approaches have also been studied extensively \cite{wang_learning_2024, sorourifar_physics-enhanced_2023, tac_data-driven_2022}. Though a neural network-based method avoids specifying a library of candidate functions, it does not readily provide a parsimonious form of the governing differential equation. These black-box models of the system dynamics are closer in spirit to operator learning---the goal is to obtain an accurate and hopefully generalizable predictive model for a particular system. The focus of this work will be on equation discovery methods such as SINDy, which claim to be interpretable by discovering sparse analytical governing equations from data.

\paragraph{} Similar to equation discovery, symbolic regression seeks to uncover symbolic mathematical equations that describe a given data set. We note, of course, that neural networks are also mathematical equations, but they are of a very different kind than the sparse equations seen in physics and engineering textbooks. Unlike equation discovery, symbolic regression uses genetic algorithms to search in a large space of mathematical functions to minimize a user-defined objective \cite{udrescu_ai_2020, cranmer_interpretable_2023, makke_interpretable_2024, schmidt_symbolic_2010}. Because this process purports to replace the human scientist in the search for physical laws, symbolic regression is seen as having wide-ranging implications for the future of science. Not surprisingly, this has generated a lot of excitement in the news \cite{wood_machine_2022, pavlus_physicist_2025, lewton_will_2022}. Symbolic regression methods have been used in climatology \cite{abdellaoui_symbolic_2021}, materials science \cite{wang_symbolic_2019}, as well as orbital mechanics \cite{lemos_rediscovering_2022}. One eminent symbolic regression researcher is optimistic that these techniques herald a new era of scientific discovery in which ML algorithms not only fit patterns in data, but also distill fundamental scientific insights \cite{cranmer_next_2024}. Interpretability is seen as a key ingredient in translating learned patterns to fundamental insights. 

\paragraph{} While researchers in these two fields extensively use the term interpretability---often upholding it as the sine qua non of genuine scientific discovery---they are hesitant to give a precise definition. Additionally, they do not utilize existing research in the field of interpretable ML \cite{rudin_interpretable_2021, molnar_interpretable_2025, murdoch_definitions_2019} to inform their arguments. One thing is clear from their discussions: the black-box nature of neural network models in some way inhibits the integration of findings from the model into the body of scientific knowledge. While interpretability may seem like a common-sense term, we will show that it can have different meanings in different contexts and that it plays a specific role in the progress of science. To help assess claims made about the scope and prospects of interpretable ML in the context of science, we believe that the research community should demand a clear-eyed approach to definitions, along with broader engagement with relevant past research and the history and philosophy of science. To this end, our contributions in this work are the following:

\begin{enumerate}
    \item We review discussions of interpretability in the equation discovery and symbolic regression literature and show that interpretability is often equated with sparsity;
    \item We survey previous work on interpretable ML and assess its relevance to applications in the physical sciences;
    \item We show that the definitions and methods of interpretable ML, as well as the informal definition of interpretability as sparsity, are inadequate for SciML;
    \item We propose an operational definition of interpretability which, as a result of being informed by the history and philosophy of science, is better-suited for SciML;
    \item We show that our definition of interpretability calls into question the possibility of scientific discoveries that are both novel and interpretable.
\end{enumerate}

The rest of this paper is organized as follows. In Section 2, we summarize discussions on interpretability within the equation discovery and symbolic regression literature, and show that the definition of interpretability is usually implicit and occasionally lacks consensus. That being said, the majority of researchers equate interpretability with the mathematical sparsity of discovered equations. In Section 3, we survey past work on interpretability in the broader ML literature and assess whether these developments apply to SciML. In Section 4, we begin by arguing that both the previous accounts of interpretability are inadequate for SciML. In response to this, we propose a novel definition of interpretability, which we believe to be more appropriate for ML research in the physical sciences. We argue for the utility of this definition by drawing from some sources and examples. In Section 5, we close with concluding remarks.


\section{The concept of interpretability in equation discovery and symbolic regression}

\subsection{Brief literature review}

\paragraph{} Interest in automating the discovery of equations from data dates back almost two decades. In \cite{bongard_automated_2007}, an early symbolic regression framework is proposed to extract nonlinear dynamical models from biological data. In the conclusion, symbolic regression is suggested as a tool to "model increasingly complex coupled phenomena \ldots a major frontier in 21st century science." In a follow-up work, symbolic regression is used to recover conserved quantities from physics, such as energy and momentum \cite{schmidt_distilling_2009}. Here, the authors state that "the concise analytical expressions that we found are amenable to human interpretation and help to reveal the physics underlying the observed phenomenon." This work represents an early introduction of the connection between simplicity/sparsity and interpretability. These methods were also used to discover implicit equations, but with less philosophical musing \cite{schmidt_symbolic_2010}.

\paragraph{} Despite these three early papers in the period of 2007-2010 and a prescient work from 2011 \cite{wang_predicting_2011}, the flurry of equation discovery research did not begin until 2016 with SINDy. Though there is no explicit discussion of interpretability in the original SINDy paper, they compare their approach to Kepler's three laws of planetary motion, and contrast it with "tools to understand static data based on statistical relationships" \cite{brunton_discovering_2016}. Though implicit, this is a clear reference to large ML models that are not interpretable. That same year, the basic SINDy algorithm was extended to partial differential equations \cite{rudy_data-driven_2016}. From the author's words, it is easy to see that their goals are scientific discovery, as opposed to the construction of black-box predictive models: "there remain many complex systems that have eluded quantitative analytic descriptions or even characterization of a suitable choice of variables \ldots we propose an alternative method to derive governing equations [for these systems]." These follow-up equation discovery works have sparked much interest in equation discovery using modern computing infrastructure. In response to concerns around the choice of variables, a subsequent work used autoencoders to carry out SINDy on learned latent variables \cite{champion_data-driven_2019}. In this work, the connection between sparsity and interpretability is explicit: "We have presented a data-driven method for discovering interpretable, low-dimensional dynamical models and their associated coordinates from high-dimensional data. The simultaneous discovery of both is critical for generating dynamical models that are sparse and hence interpretable." These core works chronicle how early on, equation discovery researchers settled on sparsity as a definition of interpretability. Note that in the forthcoming discussions of sparsity, we do not question the mathematical benefits of sparse regularization \cite{donoho_compressed_2006}. Our interests lie in arguments made about sparsity which transcend these mathematical benefits.

\paragraph{} With the groundwork laid for interpretable equation discovery in the deep learning era, many authors have refined and extended these methods. Though inspired by the idea of interpretability as sparsity, the definition of interpretability remains opaque. Using a SINDy-type algorithm, \cite{tripura_discovering_2024} shows that the Lagrangian of a dynamical system can be learned from data. The authors claim that their method is interpretable without giving a clear definition, but it is clear from context that they mean the Lagrangian has a compact, analytical form given by terms taken from a prespecified library. In \cite{lu_discovering_2022}, equation discovery methods are extended to sparse observations of the system. While they claim their method to be interpretable, they do not specify exactly what they mean. The authors write that "sparse symbolic identification approaches work directly with the governing equations of motion, which are often sparse and provide a highly interpretable representation of the dynamical system that also generalizes well." Although the authors are not explicit about this, we understand this quote to mean that generalization is a consequence of sparsity, and thus of interpretability. This conclusion is supported by additional contextual clues involving the terms "generalize" and "generalization" throughout the paper. By generalization, we mean the ability of a learned model to make accurate predictions beyond the data on which it was trained. For example, linearly elastic constitutive models calibrated from data on one structure readily generalize to other geometries. One might find evidence for the claim that interpretable models generalize better by noting that the extrapolation properties of a simple analytical expression can be understood by inspection. We will group claims about generalization/extrapolation under a definition of interpretability as "transparency," by which we mean that the behavior of the model over a wide range of inputs is easy to understand. When the model is transparent in this way, the generalization properties of the model are straightforward to assess. Interestingly, interpretability in \cite{massonis_distilling_2023} is equated with understanding the physical mechanism or meaning of a term in an equation. Using a variant of SINDy and ensuring that discovered terms come from a library of expressions with known physical meaning, they argue that the equations they discover are mechanistically interpretable. We call this the "mechanism" definition of interpretability. In \cite{wolf_interpretable_2024}, SINDy is used with dimensionality reduction to learn control laws for systems governed by partial differential equations. The authors claim that the equations learned on latent variables are interpretable, without providing any indication of how the latent space of an autoencoder can be interpreted. From context, it seems that they mean the discovered latent dynamics have a mathematical form that is simple compared to a deep neural network.

\paragraph{} Departing from the library-based methods of SINDy, the authors in \cite{desai_parsimonious_2021} hypothesize that promoting sparsity in neural networks will lead to more interpretable models. Interpretability is assessed by a recovered expression's proximity to known physical principles, such as Newton's second law. This is an idea of interpretability which makes use of both sparsity and mechanism. In \cite{ranasinghe_ginn-lp_2024}, modifications are made to a standard feed-forward neural network architecture to make its output more interpretable. Once again, interpretability is not defined, but from context, it seems to mean that the final expression comprises polynomials as opposed to less familiar functions. 

\paragraph{} A related line of work is interpretable constitutive modeling, which is a kind of equation discovery with more prior knowledge about the form of the equation. In \cite{garbrecht_interpretable_2021}, symbolic regression is used to discover nonlinear constitutive models for materials, and interpretability is defined as "transparency of model behavior [with respect to input variables]." In \cite{flaschel_unsupervised_2021}, sparse regression on a dictionary of candidate functions is used to build hyperelastic constitutive models. So-called "uninterpretable" neural network-based models are critiqued based on limiting "the insight that they can bring towards the physical understanding of the material behavior." This vague notion of interpretability suggests a mix of transparency, sparsity, and mechanism. Finally, in \cite{fuhg_extreme_2023}, sparse neural networks are used to discover hyperelastic constitutive laws that are sufficiently simple to be written down analytically. Interpretability is identified with sparsity, and sparsity is seen as a boon as it gives the user "knowledge about [the model's] extrapolation behavior." 

\paragraph{} Discussions of interpretability in recent symbolic regression papers closely follow those of equation discovery. In \cite{udrescu_ai_2020}, a large number of algebraic equations from physics are recovered with symbolic regression. Though there is no explicit invocation of interpretability, a comparison is again made between symbolic regression and Kepler's discoveries. We take this to be the sparsity definition of interpretability. In \cite{cranmer_interpretable_2023}, the author reviews a particular implementation of a symbolic regression tool, also citing Kepler as the paradigmatic example of fitting symbolic expressions to data. Like other works, there is a clear emphasis on the importance of interpretability, but little clarity about its meaning or the specific role that it plays in scientific discovery. In \cite{lemos_rediscovering_2022}, the symbolic form of the gravitational force is discovered from data on orbital trajectories. Here, the authors state the following about interpretability: "describing physical phenomena with compact symbolic formulations supports scientific interpretation, and can interface with existing symbolically defined physical theories." No definition of interpretability is given, but a connection between interpretation and sparsity is made. Furthermore, we see allusion to the idea that the findings of ML models and traditional physical theories are cast in two different languages. Though a clear definition is lacking, the quote from \cite{lemos_rediscovering_2022} hints at a deeper motivation for interpretability in SciML: the findings of a neural network are illegible to the scientist and thus cannot be drawn into the existing body of scientific knowledge.

\paragraph{} Another example of interpretability defined with sparsity and transparency is \cite{wang_symbolic_2019}. Some other symbolic regression papers that implicitly equate interpretability and sparsity are \cite{abdellaoui_symbolic_2021, li_advancing_2024, wang_discovering_2024}. In a review on symbolic regression, one work provides an explicit definition of interpretability: "a model is interpretable if the relationship between the input and output of the model can be logically or mathematically traced succinctly. In other words, learnable models are interpretable if expressed as mathematical equations" \cite{makke_interpretable_2024}. This is a direct reference to sparsity and transparency. Finally, we note that some approaches do not rely on genetic programming to solve the symbolic regression problem. An example is \cite{guimera_bayesian_2020}, where a Bayesian approach is taken. In this work, a similarly vague notion of interpretability is employed. Once again, context clues suggest that interpretability is taken to be the result of simple mathematical expressions.

\begin{table}[h]
    \centering
    \begin{tabular}{|c|c|c|c|}
        \hline
          \textbf{Author(s)} & \textbf{Sparsity} & \textbf{Transparency} & \textbf{Mechanism} \\ \hline
          Bongard \& Lipson \cite{bongard_automated_2007} & \checkmark   & \xmark   & \xmark   \\ \hline
          Lipson \& Schmidt \cite{schmidt_distilling_2009} & \checkmark & \xmark & \xmark \\ \hline
          Brunton et al. \cite{brunton_discovering_2016} & \checkmark & \xmark & \xmark \\ \hline
          Champion et al. \cite{champion_data-driven_2019} & \checkmark & \xmark & \xmark \\ \hline
          Tripura \& Chakraborty \cite{tripura_discovering_2024} & \checkmark & \checkmark & \xmark \\ \hline
          Lu et al. \cite{lu_discovering_2022} & \checkmark & \checkmark & \xmark \\ \hline
          Desai \& Strachan \cite{desai_parsimonious_2021} & \checkmark & \xmark & \checkmark \\ \hline
          Massonis et al. \cite{massonis_distilling_2023} & \checkmark & \checkmark & \checkmark \\ \hline
          Garbrecht et al. \cite{garbrecht_interpretable_2021} &  \checkmark & \checkmark  & \xmark \\ \hline
          Flascehl et al. \cite{flaschel_unsupervised_2021} & \checkmark & \checkmark & \xmark \\ \hline
          Fuhg et al. \cite{fuhg_extreme_2023} & \checkmark & \checkmark  & \xmark \\ \hline
          Udrescu \& Tegmark \cite{udrescu_ai_2020} & \checkmark & \xmark & \xmark \\ \hline
          Cranmer \cite{cranmer_interpretable_2023} & \checkmark & \xmark & \xmark \\ \hline
          Wang et al. \cite{wang_symbolic_2019} & \checkmark & \checkmark & \xmark \\ \hline
          Makke \& Chawla \cite{makke_interpretable_2024} & \checkmark & \checkmark & \xmark \\ \hline
          Guimera et al. \cite{guimera_bayesian_2020} & \checkmark & \xmark & \xmark \\ \hline
          
          \hline
    \end{tabular}
    \caption{Summarizing the notions of interpretability employed by some of the prior works in equation discovery and symbolic regression that we have reviewed.}
    \label{tab:summary}
\end{table}

\paragraph{} Though the importance of interpretability for doing science is widely acknowledged, the reader is often left to infer the meaning of the term from context, and sometimes to negotiate between multiple definitions. See Table \ref{tab:summary} for a summary of the three primary notions of interpretability we have identified: sparsity, transparency, and mechanism. In line with the works that we have reviewed, by sparsity we mean compact, mathematical expressions that comprise familiar algebraic and transcendental functions of quantities of interest and their derivatives (in the case of partial differential equations). We use the term transparency to refer to the ability of the analyst to look at an expression and easily infer its behavior over a wide range of inputs. Mechanism has to do with understanding the physical meaning of an expression. As Table \ref{tab:summary} shows, the vast majority of aforementioned papers equate interpretability with sparsity.

\subsection{Discussion}

\paragraph{} Our literature review suggests that the SciML community has converged on the concept of interpretability to differentiate between neural network-based models and the analytical equations familiar from the physical sciences. Unlike traditional ML researchers, scientists operate under the assumption that underneath the apparent heterogeneity of their data lies a small set of fundamental principles packaged concisely in short, analytical expressions. As the epochal discoveries of Kepler, Newton, Maxwell, and Einstein have all corroborated this belief, scientists have good reasons to trust this idea. We note that this belief in reduction to a small set of underlying principles is an article of faith with an origin in Greek philosophy, which has survived because of its utility in describing many natural phenomena \cite{feyerabend_tyranny_2013, toulmin_fabric_1999}. This belief in what we call "sparse reductionism" is the justification for emphasizing interpretability in SciML. For example, it need not be that the mechanism of a classification algorithm in computer vision is human understandable (see the discussion on rule-based systems in the introduction), but scientists routinely assume that natural phenomena ought to reduce to a comparatively simple set of mechanisms. Ideally, ML models could be used to discover this sparse set of fundamental principles, as opposed to their comparatively crude function as over-parameterized regression models. We note that the belief in sparse reductionism---and the corresponding desire to pull phenomena back to underlying principles---also accords with Philip Kitcher's notion of explanation as unification \cite{kitcher_explanatory_1981}. According to Kitcher, good scientific explanations seek to describe as much of nature as possible with a small set of fundamental laws. The laws themselves are treated as axiomatic and therefore cannot be explained. We want to flag that, despite its apparent power, sparse reductionism does not apply to all systems. For example, the field of complexity science grapples with natural systems that stubbornly resist distillation to a simple set of underlying principles \cite{holland_hidden_1996, morowitz_emergence_2002, weaver_science_1948}. Though we will adopt the SciML community's assumption that systems reduce to a small set of principles, we suggest that this is a prejudice that need not hold in all cases. More important for our purposes, we will discuss how this belief in simple mechanisms (and thus sparse equations) can become tangled with interpretation. We propose that sparse reductionism and interpretation are best thought of as overlapping but not equivalent notions.

\paragraph{} Though it is not often made explicit in the literature, we propose that the real concern with interpretability in SciML is based on the desire to integrate findings from data into the body of scientific knowledge. Philosopher of science W.V. Quine imagines scientific knowledge "as a field of force whose boundary conditions are experience" \cite{quine_two_1951}. Doing science with black-box ML models trained on large data sets (experience, in Quine's words) thwarts this vision, as the findings of an ML model are not cast in the same language as the body of scientific knowledge. The language of the physical sciences is written in terms of concise mathematical expressions, usually making use of algebraic and transcendental functions. Standard ML models are verbose in comparison, containing hundreds, thousands, or millions of parameters that construct exotic mathematical relations. How can the "force field" of scientific knowledge accommodate itself to boundary conditions it does not understand? Our review suggests that the underlying idea of interpretability, as it is encountered in the literature, is to cast the findings of ML models in the language of traditional science. It is clear in many cases that researchers believe that the findings of a data-driven model, which outputs simple mathematical expressions, can be integrated into the body of scientific knowledge. This suggests the sparsity definition of interpretation and that interpretation is valuable because it allows knowledge to accumulate. This argument raises some concerns, however. Are all simple mathematical expressions also simple to interpret? How exactly is a discovery integrated into the storehouse of scientific knowledge? To clarify the discussion in the literature, we must address the following two questions: 


\begin{enumerate}
    \item What is interpretability in the context of science? 
    \item How does interpretability advance scientific knowledge?
\end{enumerate}

\paragraph{} We will argue that sparsity is directionally correct as a definition of interpretability for science, but insufficiently specific. We believe that sparsity alone does not guarantee that an equation is interpretable. We note that agreement on the goal of integrating insights from data into the body of scientific knowledge does not answer the question of how this is accomplished (question 2). Many authors have unwittingly proposed some answers in their works, but these questions have not been tackled head-on. Before proceeding to provide our answers to these questions, we must verify that the broader field of interpretable ML does not already hold the key to either of the two questions we have posed.



\section{Interpretable and explainable ML}

\subsection{Brief literature review}

\paragraph{} Traditional ML research uses data-driven algorithms, often involving neural networks, for various classification and regression tasks. This includes but is not limited to image classification \cite{alom_history_2018}, autonomous vehicles \cite{ks_pdf_nodate}, speech recognition \cite{ahlawat_automatic_2025}, drug discovery \cite{yang_pdf_nodate}, policing \cite{sarzaeim_systematic_2023}, approving loans \cite{uddin_ensemble_2023}, and recommendation algorithms \cite{roy_systematic_2022}. Many of the motivations for pursuing interpretability in the broader ML community are quite clear: transparency, safety, ethics, trust, and debugging. These concerns have given rise to the subfields of "interpretable ML" and "explainable AI" (XAI). Interpretable ML seeks to build data-driven models which are interpretable by construction (such as decision trees), whereas XAI develops methods to extract post-hoc interpretations from black-box models \cite{garouani_investigating_2024}. Because both approaches aim to understand the behavior of a data-driven model, we do not carefully distinguish between the two in our review. In \cite{boon_epistemological_2024}, over-reliance of physicians on data-driven models is criticized as obscuring less quantifiable forms of information. Here, transparency lays bare the fact that the models only make use of partial information. In \cite{bereska_mechanistic_2024}, understanding the mechanisms of how a model makes decisions is upheld as a crucial step in assessing its safety. Other authors champion interpretability for models making consequential decisions as a safeguard against discrimination and bias \cite{goodman_european_2017}. Interpretability is cited as a necessary prerequisite for humans to trust data-driven models \cite{kim_interactive_2015}. Interpretability is also seen as an aid in debugging and improving a model, as a model that learns to perform on the training data by learning the wrong relationship can be identified and corrected \cite{molnar_interpretable_2025}. Most authors in the field of interpretable ML agree with a subset of this basic list of motivations for interpretability \cite{doshi-velez_towards_2017, lipton_mythos_2017, murdoch_definitions_2019, marcinkevics_interpretable_2023}.

\paragraph{} The above discussion shows that the reasons for pursuing interpretability are laid out in the interpretable ML literature. What about our first question regarding the definition of interpretability? In \cite{doshi-velez_towards_2017}, the authors state that: "in the context of ML systems, we define interpretability as the ability to explain or present in understandable terms to a human." Another definition of interpretability reads: "an interpretable model is constrained, following a domain-specific set of constraints that make reasoning processes understandable" \cite{rudin_interpretable_2021}. In \cite{lipton_mythos_2017}, a taxonomy of properties of interpretable models is outlined, whereby interpretation splits into two broad categories. On one hand, some models are interpretable by construction, and on the other, some models require post-hoc processing to extract interpretations. This is a common point of discussion in works which focus on methodologies for interpretation \cite{molnar_interpretable_2025, marcinkevics_interpretable_2023}. Noting vagueness around the definition of interpretability in the literature, another set of authors propose a definition of interpretation as "extraction of relevant knowledge from a machine-learning model concerning relationships either contained in data or learned by the model" \cite{murdoch_definitions_2019}. Last but not least, the authors in \cite{gilpin_explaining_2019} define interpretability as describing "the internals of a system in a way that is understandable to humans," and also suggest that a model should be able to answer "why" questions posed to it to be understandable. An example question would be: "Why does this particular input lead to that particular output?", where an answer that simply reads out the model parameters would not be satisfactory. Rather, an understandable answer must make use of concepts and/or references that are familiar and meaningful to the relevant audience. It is clear that these definitions of interpretability center around an intuitive notion of understanding, which we believe to be satisfactorily captured by the ability to answer "why" questions. An interpretable computer vision model informs the user that an image is classified as a dog because it has two pointy ears and fur. An interpretable autonomy algorithm for a driverless car might communicate that it braked because there was an object on the road. An interpretable recommendation algorithm can report that a given recommendation is a consequence of past viewing history. There is a certain "if X, then Y" logic to these explanations, where X is some familiar feature of the system under study and Y is the outcome. Thus, the question arises: Is a definition of interpretation based on an intuitive causal understanding appropriate for science?


\subsection{Discussion}

\paragraph{} We now explore whether the motivations for and definitions of interpretability in the interpretable ML literature can be ported to the problems of SciML. The desiderata of transparency, safety, ethics, trust, and debugging are certainly of interest for technologies that use scientific insights distilled from data. However, they are not paramount in the research of a scientist who is attempting to discover the fundamental rules that govern a system. In other words, equations thought to describe strictly physical regularities---even those informing the design of crucial infrastructure---are not scrutinized based on safety, ethics, trust, etc. Questions of this sort directly involve human judgment and arise only in the application of the basic science, not in the formulation of the basic science. Thus, we see that the field of interpretable ML will not answer the second question we posed on the importance of interpretability in science. Given that the goal of the scientist is to discover fundamental laws from data and integrate these laws into the body of scientific knowledge, the goals of the interpretable ML community, which revolve primarily around effective integration of ML-based technologies into society, will differ from those of the SciML community.

\paragraph{} Our review of various definitions of interpretability showed that researchers agree that an intuitive causal understanding of a model's input-output relation is tantamount to interpretation. This is analogous to the sparsity argument we encountered in the fields of equation discovery and symbolic regression---sparse models have transparent input-output relations and should therefore be interpretable. We note that in the context of sparse discovered differential equations, this straightforward input-output relation is between the input and the relevant derivatives of the state, not between the input and the state itself. This observation does not change our claim that there is a basic similarity between the intuitive-causal definition of interpretability and the sparsity definition from SciML. Thus, the explicit definitions of the interpretable ML community and the informal consensus of scientific researchers seem to agree that sparsity is the path to interpretability. Does this provide an answer to our question: "What is interpretability in the context of science?" No, we propose. This is because, for the scientist concerned with understanding the mechanisms that underlie phenomena, sparsity alone does not guarantee that an equation is interpretable. As we will show in the following section, mathematical sparsity of an equation has no clear way of interfacing with sparse reductionism, which is the idea that the dynamics of many systems can be explained with a small set of underlying principles. A sparse equation may be just as isolated from the body of scientific knowledge as a neural network model, and we argue that an equation describing a physical system that is disconnected from fundamental principles is not interpretable. Our illustrations of this point in the next section will assist in rectifying the definition of interpretation for SciML.




\section{The definition and importance of interpretability}

\subsection{Failure of the sparsity definition}

\paragraph{} Because we have yet to propose a formal definition of interpretability, our task is to show that the mathematical sparsity of an equation can lead to contradictions with the common-sense usage of interpretation. This tension will be the impetus for establishing a novel definition of interpretation for SciML. To be clear, because our definition of interpretation is tied specifically to SciML, the subject doing the interpretation is taken to be a scientist or researcher trained in the methods and theories of the relevant field. Now, consider an elastic cantilevered beam with an applied end force. The displacement at the end of the beam is called $u$, and the force has magnitude $F$. We are interested in understanding the time-independent mechanics of this system, so we set up an experiment and record the end displacement as a function of the applied force. We might use any manner of symbolic or sparse regression techniques to obtain an expression such as the following:

\begin{equation}\label{simple}
    u = \lambda_1 F - \lambda_2 F^2,
\end{equation}

\noindent where $\lambda_1, \lambda_2 >0$ are the empirical coefficients. We note that this example is a thought experiment, not a fit obtained from a real data set. But in practice, Eq. \eqref{simple} could be recovered by doing sparse regression on data with a basis given by the monomials $\{ F^i \}_{i=1}^M$. This would be an example of a dictionary-based method, such as SINDy \cite{brunton_discovering_2016}. This equation is certainly sparse, and its generalization properties are plain to see. Notice that it predicts that at some point, increasing the force decreases the displacement. The analyst might conclude on physical grounds that this is an artifact of the poor extrapolation properties of the model. By having skepticism about using the model in this extrapolation range, the analyst might prevent a problem that would have occurred with a black-box model. But is this model interpretable? The practitioner of solid mechanics will be quick to identify the first term as a linear elastic response and the quadratic term as introducing concavity to capture a geometrically nonlinear stiffening effect. But what of scientific interest has been learned from our interpretable sparse model? We have learned that we already knew a lot about the problem we set out to solve. In our thought experiment, an analyst with sufficient prior knowledge to interpret these two terms as a combination of linear elasticity and a geometrically nonlinear effect has little fundamental scientific interest to learn from this data-driven model. Of course, making use of prior knowledge to fit parametric models to data is an extremely common and important task in science. But this is the domain of inverse problems---which give insight into the presence or absence of known mechanisms in a system, to be sure---but it is not the discovery of novel scientific laws. Take the same model of Eq. \eqref{simple} and alter the meaning of $u$ and $F$. Say that $F$ represents the number of hours spent studying for a test, and $u$ represents performance relative to the mean. Is it clear what the two terms represent anymore? Is the mathematical form helpful in demarcating a point beyond which the model ceases to make sense? In other words, is it clear that sparsity is a boon in assessing the generalization properties? Does the sparsity of the mathematical form of the model guarantee its interpretability? With this example, we note that it is often the case that simple mechanisms yield sparse equations, but moving in the other direction is not so simple---it is not clear that there is any process to back out mechanisms from simple equations. For example, the statistical thermodynamics of gases provides one example where sparse equations are the consequence of complicated underlying mechanisms. The sparse equations of ideal gases were known for decades, \cite{wang_brief_2011}, yet this did not shed light on the molecular mechanisms that are required to make sense of their behavior in terms of classical and quantum mechanics.

\begin{figure}[hbt!]
\centering
\includegraphics[width=0.9\textwidth]{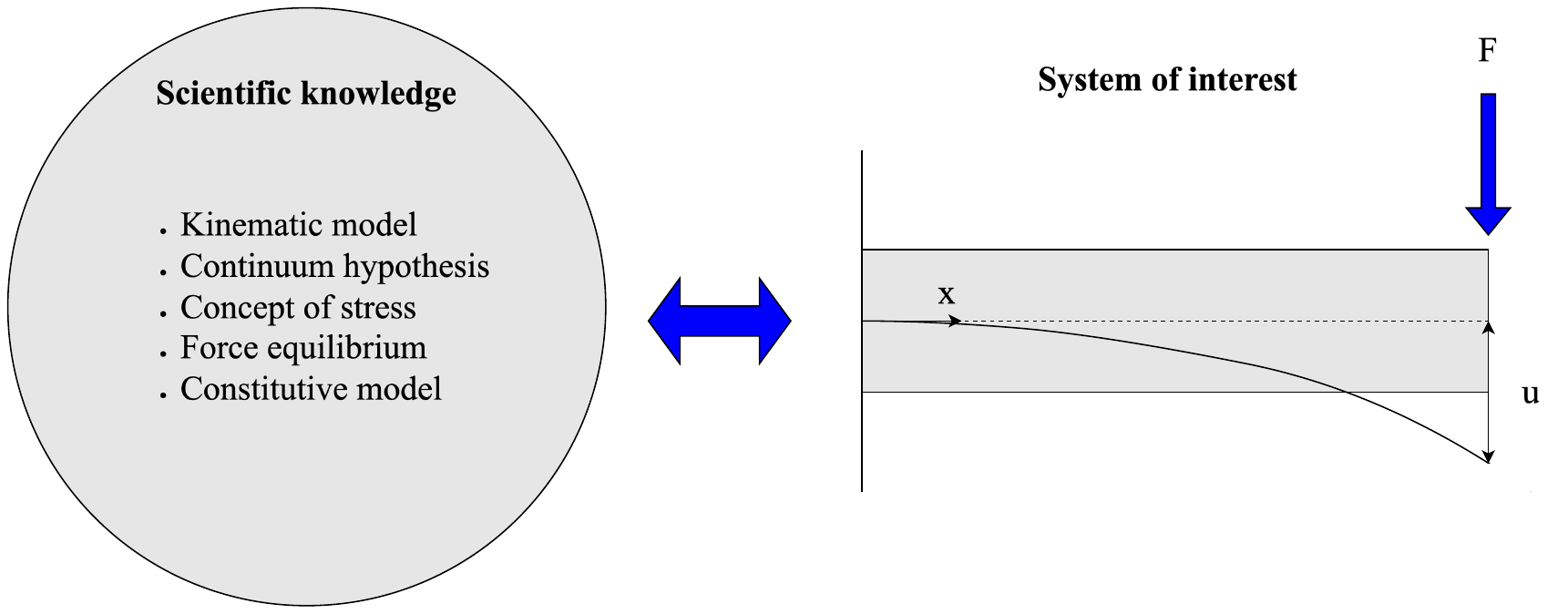}
\caption{A thought experiment to show that sparsity does not guarantee interpretability in the absence of prior knowledge. In science, we propose that the goal of interpretation is to make a connection between a particular system of interest and the relevant set of underlying principles.}
\label{beam}
\end{figure}

\begin{figure}[hbt!]
\centering
\includegraphics[width=0.7\textwidth]{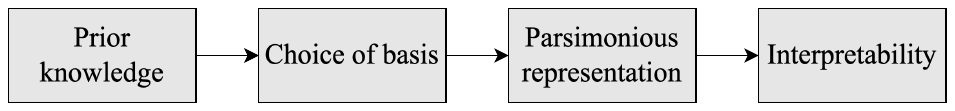}
\caption{Interpretable equations are guaranteed when prior knowledge of mechanisms restricts the choice of basis used in the regression problem.}
\label{prior knowledge}
\end{figure}

\paragraph{} With the beam problem example, the point we wish to raise is that in science, prior knowledge must be brought to bear on an expression to interpret it. A machine learning researcher interested in interpretability may be content with a model that is accurate and predictable in its generalization properties. In having transparent input-output relations, sparse models answer the "why?" question of such a researcher---the answer that $u$ begins to decrease because $F > \lambda_1/2\lambda_2$ is sufficient in this setting. However, as a result of faith in sparse reductionism, the scientist asks "why?" on a deeper level, and thus demands a different type of interpretation. This is because interpretation in science revolves around connecting an expression to an underlying set of fundamental principles, as shown in Figure \ref{beam}. The beam example shows that sparsity is not sufficient for interpretability, and a subsequent example will show it is not even necessary. We note, however, that one account of interpretation in the equation discovery and symbolic regression literature is that of Figure \ref{prior knowledge}. If every candidate function in the basis used to represent the unknown equation is associated with a particular physical mechanism, then the discovered equation will be interpretable. As discussed above, this limits interpretation to problems where the underlying mechanisms are already known.

\paragraph{} We now consider more realistic examples, the first of which is taken directly from the literature. We show two expressions of varying mathematical complexity, and argue that the more sparse equation is less interpretable. By mathematical complexity, we mean a combination of the length of the expression and the kinds of functions it comprises. For example, though the expression $x+y$ has two terms, we take the expression $\sin(\tanh(xy))$ to be more complex given the composition of nonlinear functions. We do not attempt to quantify this complexity, as we believe an intuitive notion of mathematical complexity suffices for our purposes. Referring to \cite{fuhg_extreme_2023}, an example of a hyperelastic strain energy density discovered from data using a sparse neural network is

\begin{equation}\label{psi}
    \hat \Psi = 0.398J + 3.095 \log\Big( ( 1 + e^{-1.356I_2})^{1.314} (e^{0.755I_1} + 1 )^{0.515} (e^{0.135 I_1 - 0.319 I_2 - 0.329J} + 1)^{1.874} + 1 \Big)- 6.686.
\end{equation}

This expression makes use of the following definitions and identities, which will be familiar to the practitioner of hyperelastic constitutive modeling:

\begin{gather*}
    \mathbf{F} = \mathbf{I} + \pd{\mathbf{u}}{\mathbf{X}}, \\
    J = \text{det} \mathbf{F}, \\
    I_1 = \text{tr}(\mathbf{F}^T \mathbf{F}),\\
    I_2 = \frac{1}{2}\Big(  \text{tr} (\mathbf{F}^T \mathbf{F})^2 - \text{tr}((\mathbf{F}^T \mathbf{F})^2)\Big).
\end{gather*}

Here, $\mathbf{u}$ is the displacement and $\mathbf{X}$ is the spatial coordinate in the undeformed configuration of the structure. We include the definitions of the deformation gradient $\mathbf{F}$, its determinant $J$, and the invariants $I_1$ and $I_2$ for completeness, but the details of hyperelastic constitutive modeling are not our focus. According to \cite{fuhg_extreme_2023}, the interpretability of Eq. \eqref{psi} is justified on the grounds of its sparsity. A standard feed-forward neural network might require an entire page to write out, as opposed to a single line. Though this expression is sparse relative to a standard neural network, it is by no means simple. We will compare the interpretability of this expression to an equation obtained from a thought experiment involving a different physical system. Imagine we have data on a distributed system which we believe involves the advection, diffusion, and reaction of a scalar quantity $c(x,t)$ in one spatial dimension. Following the methods of \cite{rudy_data-driven_2016}, we write the governing equation as 

\begin{equation} \label{SINDysetup}
    \pd{c}{t} = \mathcal{N}\qty( c,\pd{c}{x}, \frac{\partial^2 c}{\partial x^2},\dots; \boldsymbol \lambda),
\end{equation}

\noindent where $\boldsymbol \lambda$ is a vector of coefficients used to scale terms in a library. This library of terms is taken to represent the right-hand side of the equation, and sparse regression techniques can be used to recover the relevant terms from the library per \cite{rudy_data-driven_2016}. In this thought experiment, we carry out the analysis and presume to recover the following equation:

\begin{equation}\label{SINDyreaction}
    \pd{c}{t} + \lambda_1  \pd{c}{x} - \lambda_2 \frac{\partial^2 c}{\partial x^2} - \lambda_3 c + \lambda_4 \Big | \pd{c}{x}\Big| c  = 0.
\end{equation}

We have recovered a sparse equation and thus, according to many of the authors reviewed above, an interpretable one. For the sake of argument, let us imagine that our recovered model is even predictive on unseen data. This equation is more sparse than the hyperelastic energy density in Eq. \eqref{psi}. By all metrics employed in the literature, we have successfully solved our problem. But is this equation more interpretable than the strain energy? Is it interpretable at all? To explore this, we can assign meanings to the different terms:

\begin{equation}\label{SINDyinterpret}
    \underbrace{\pd{c}{t}}_{\text{time evolution}} + \underbrace{ \lambda_1  \pd{c}{x}}_{\text{advection}} - \underbrace{\lambda_2 \frac{\partial^2 c}{\partial x^2}}_{\text{diffusion}} - \underbrace{\lambda_3 c}_{\text{reaction}} + \underbrace{\lambda_4  \Big | \pd{c}{x}\Big| c }_{?} = 0.
\end{equation}

As before, we have drawn from prior knowledge of the system to interpret terms. In this case, advection, diffusion, and reaction are particular mechanisms of mass transport. Three of the four terms are familiar and can be interpreted this way. But the fourth term, though having a simple mathematical form, is unfamiliar. The reader may object that this term is unfamiliar because it is not physical. To this, we respond: if the interpretation of an equation relies on an {\it a priori} ability to distinguish between meaningless and meaningful terms, we have ruled out the possibility of scientific discovery. In other words, if we grant that scientific discovery---as opposed to inverse problems, in which the general form of the missing term(s) is known---revolves around encounters with the unfamiliar, then prior knowledge is absent and interpretation is not possible. In contrast with the unfamiliar term in Eq. \eqref{SINDyinterpret}, and despite its mathematical complexity, the interpretation of the strain energy density is abundantly clear. The strain energy density represents the nature of energy storage in a material undergoing reversible deformations. That it has a rather esoteric mathematical form does not complicate its interpretation as encoding a dependence between deformation and energy. We propose that this is true even if the strain energy was written as a non-sparse neural network of the most general variety. Though its physical interpretation is clear, the mathematical dependence of the strain energy density on components of the deformation gradient in Eq. \eqref{psi} is complex, despite the sparsification of its neural network representation. This expression is not amenable to interpretation by inspection, and we suggest that it does not take a great deal of mathematical complexity for this to be the case. On a more fundamental level, we reject the equivalence drawn between interpretability and generalization, as the problem of induction states that the ability of a model to generalize is always in question, independent of its complexity \cite{henderson_problem_2024}. In summary, the strain energy density is interpretable because it has a meaning that connects it back to fundamental concepts from continuum mechanics.


\paragraph{} Whereas the complex mathematical form of the strain energy does not preclude interpretation, the sparsity of the unfamiliar discovered term in the advection-diffusion-reaction system does little to facilitate interpretation. We reiterate that we still have not provided a formal definition of interpretation, though the idea has begun to crystallize in the preceding discussions. We also note that we are not looking for a "true" definition of interpretation, simply one that accords with intuition and that is useful to SciML practitioners. Given that interpretability is a desideratum, the usefulness of the definition can be gauged by the extent to which it focuses efforts on questions that lead to the accumulation of scientific knowledge. With these examples, we are trying to surface contradictions between sparsity and interpretation, using their common-sense meaning for scientists. The examples also demonstrate that sparsity needs to provide a satisfactory connection to pre-existing scientific knowledge. We \textit{interpret} the strain energy density as encoding energy storage mechanisms. We \textit{interpret} terms in fluid systems as representing particular mechanisms of mass transport. We \textit{interpret} terms in the governing equations of continuum mechanics as being mechanisms of force generation. As scientists, we do not interpret abstract mathematical relations with no connection to physical principles. Though this is entirely contrary to the intentions of the authors, the recurring references to Kepler and his famous laws of planetary motion \cite{cranmer_interpretable_2023, udrescu_ai_2020, champion_data-driven_2019, brunton_discovering_2016} perfectly exemplify the potential disparity between sparsity and interpretability. Published in 1619, Kepler's three laws of planetary motion are:

\begin{itemize}
    \item Planets move in elliptical orbits with the sun as one of the two foci;
    \item At all positions of a planet's orbit, a line drawn from a planet to the sun sweeps out equal areas over a given unit of time;
    \item The square of a planet's orbital period is proportional to the cube of the semi-major axis of the orbit.
\end{itemize}

Of course, these laws can also be cast as simple mathematical expressions. Though sparse, Kepler's laws were by no means interpretable. At the time, Kepler had no idea why these laws held. Despite their mathematical sparsity, Kepler's laws only became interpretable when Newton showed 70 years later that they were derivable within the framework of his mechanics and the inverse square law of gravity \cite{westfall_life_1994}. To put this in perspective, Kepler was not even capable of using the concept of gravitational force as a tool for reasoning about these results, given that the modern version of this concept was unknown to him. Though compact mathematical expressions with superb generalization properties are clearly of scientific interest, compactness on its own is no guarantee of interpretability. 

\paragraph{} We see that interpretations are not beholden to sparsity---they are beholden to prior knowledge. When a so-called interpretable equation discovery method produces a term that we do not recognize from experience, it will not be interpretable, regardless of the sparsity of its mathematical form. But, as the example of Kepler shows, sparse expressions are certainly of scientific interest. Expressions of this sort can be used to motivate a search for underlying mechanisms or to help verify a new theory. These are separate notions from interpretability, however. In response to the deficiencies we have diagnosed in the sparsity definition of interpretation, we can now formalize our ideas about the role of prior knowledge into a definition of interpretation that is appropriate for SciML.

\subsection{Novel definition of interpretability}


\paragraph{Definition} In SciML, a learned model is interpretable when it can either be derived from basic physical principles or it represents an empirical component of a model derived from basic physical principles.

\paragraph{} We note that our definition prohibits the physical principles themselves (Newton's laws, conservation of energy, etc.) from being interpreted. This is a standard move in the philosophy of science, as these principles are the basic tools with which scientists reason, and it is not clear what one would appeal to to reason about these tools. For example, in \cite{carnap_philosophical_1966}, Carnap chronicles how asking "why" questions about fundamental laws has disappeared from science. Interpreting fundamental principles has the flavor of trying to derive axioms of a system from within that system, a task which is ill-founded. Note that, at the time of its inception, Newton worried that his concept of force would be criticized as metaphysical and was careful to make no claims about its ontological status \cite{westfall_life_1994}. The scientist Christian Huygens found the principle of gravitational attraction "absurd," and Leibniz felt compelled to make sense of it in terms of a fluid-like substance he called "aether." In practice, the scientist's hard-earned sense of familiarity with concepts such as force and energy takes the place of interpretation. This framing of interpretation bears resemblance to the deductive-nomological model of explanation introduced by Carl Hempel \cite{hempel_explanation_2001}. Hempel says that science answers "why" questions "by subsuming the uniformities under more inclusive laws, and eventually under comprehensive theories." This notion of explanation accords with our approach to interpretation for SciML.

\paragraph{} The concept of "basic physical principles" needs further clarification. By a standard reductionist account of science, the physics of elementary particles would be the basic principles underlying all fields. Yet, as Philip Anderson says, "biology is not just applied chemistry" \cite{anderson_more_1972}. Reductionism notwithstanding, it is a fact that, in practice, different fields have different principles that they treat as basic. For an engineer, classical thermodynamics may be treated as basic, whereas it is a consequence of deeper underlying theories for the statistical physicist. Newtonian mechanics is treated as a basic physical principle for analysis of terrestrial phenomena, though it can be derived from general relativity. Thus, the laws that are taken as basic are field-dependent. In SciML, most systems of interest can be treated in light of conservation of mass, momentum, and energy. These conservation laws provide a reasonable sense of what constitutes basic physical principles in our context.

\paragraph{} An immediate consequence of this definition is that revolutionary scientific discoveries---in other words, the discovery of the basic principles themselves---are not interpretable. This limits interpretable equation discovery to what Thomas Kuhn calls "normal science," meaning that scientists work on narrowly defined technical problems as opposed to speculative exploration of fundamental principles \cite{kuhn_structure_2012}. As we have mentioned, an example of this in contemporary scientific research is inverse problems, where missing parameters and/or terms of an equation are estimated using data. We believe the tension between interpretability and revolutionary discovery to be under-appreciated in the literature, so we see our surfacing of this tension as a worthwhile contribution.

\paragraph{} Our definition equates interpretability with mechanistic understanding, or the "meaning" behind the math. We believe this is closer to the scientist's intuitive definition of interpretability. Advection is interpretable because it can be derived from mass conservation in an Eulerian frame. The exotic source term found in Eq. \eqref{SINDyreaction} cannot be interpreted because the story of its origin cannot be reverse-engineered. Or, to be more precise, it is not interpretable \textit{unless} this story can be reverse-engineered. An equation becomes interpretable when it can be derived from basic principles. Similarly, all scientific models have empirical components, and by virtue of existing in the theoretical framework of the model, these empirical components have a known meaning. Newton's gravitational constant scales the force between planets. The electric permittivity scales the force that two charges exert on each other in free space. The relationship between force and deformation in a material is captured with a constitutive law. Thus, empirical expressions of this sort have physical meaning even though they are not derived. Thus, a hyperelastic strain energy density is interpretable. We remark that treating a constitutive relation of this sort as an empirical component is also field-dependent, as, in principle, the constitutive behavior of a material can be derived from its molecular structure. Of course, this would require underlying theories with their own empirical components, such as the charge and mass of electrons \cite{irving_statistical_1950}. In engineering solid mechanics, this approach is rarely adopted, and empirical constitutive relations are interpreted as quantifying material-specific energy storage, or as stress-strain relations. In SciML, empirical components of models are primarily constitutive relations. We grant that to a statistical physicist, a data-driven constitutive relation may appear to be an uninterpretable model, thus problematizing the division between empirical components and uninterpretable models. This is no matter---just as basic principles are field-dependent, so are the components of a model which a scientist treats as empirical.

\begin{figure}[hbt!]
\centering
\includegraphics[width=0.6\textwidth]{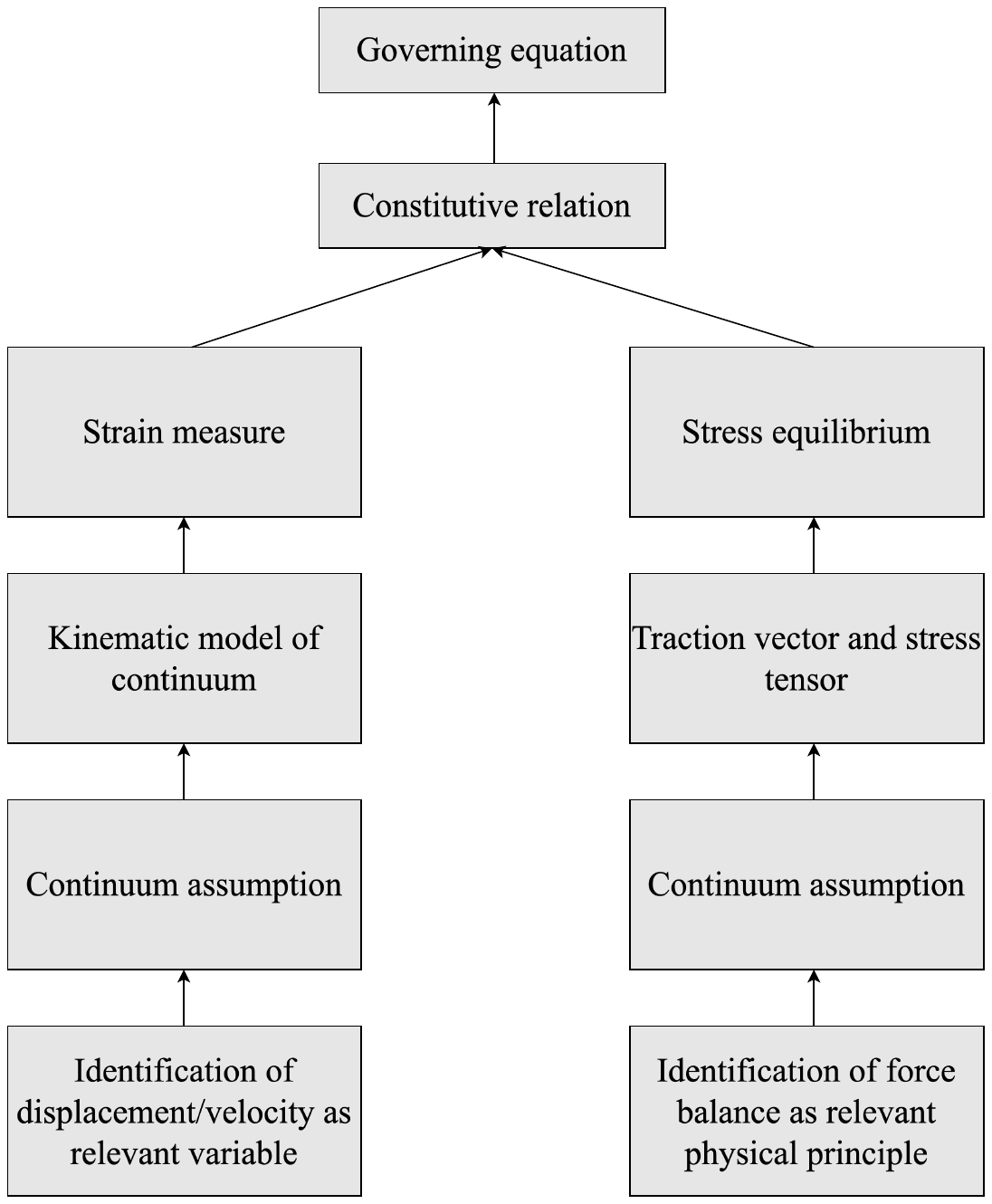}
\caption{The path to deriving the governing equations for the mechanics of a continuum.}
\label{continuum}
\end{figure}

\paragraph{} Note that nowhere does sparsity show up in this definition of interpretability. For example, Figure \ref{continuum} outlines the theoretical machinery behind continuum mechanics. The final governing equation that emerges from this story may be sparse, but is it possible to interpret it unless one knows the steps required to derive it? Our definition suggests that interpretation happens primarily in the steps that precede the governing equation. This raises the question: is there any role for sparsity in equation discovery? In the next section, we will address this and the question of the importance of interpretability for science.

\subsection{Discussion}

\paragraph{} By our definition, the sparsity of an equation does not guarantee its interpretability. But we believe methods that encourage sparsity in recovering governing equations from data are still valuable, even beyond the mathematical benefits of regularization. To see this, we will again consider the dynamics of a concentration field $c(x,t)$. In this example, there will only be advection and diffusion effects. We assume that we have two different learned representations of the system dynamics. The first is a representation we obtain from neural ordinary differential equations \cite{chen_neural_2019}, where the system dynamics are represented by a deep neural network that takes in the state and its derivatives. This has the same form as Eq. \eqref{SINDysetup}, except that the vector  $\boldsymbol \lambda$ represents the weights and biases of the neural network. The second representation is a form that is recovered from sparse regression with a suitable library of candidate functions using SINDy-type methods. An example of the latter representation reads

\begin{equation}\label{SINDy}
    \pd{c}{t} +  \lambda_1  \pd{c}{x} - \lambda_2\frac{\partial^2 c}{\partial x^2} = 0.
\end{equation}

Let us assume that we have no prior knowledge of the physical meaning of the advection and diffusion terms in Eq. \eqref{SINDy}. By our definition, these two terms are not interpretable, even though they are sparse. Thus, both the SINDy and neural ODE representations are not interpretable in the absence of prior knowledge. The question is: in the setting of scientific discovery, is there an advantage to the sparse but uninterpretable (given our state of knowledge) Eq. \eqref{SINDy} over the non-sparse representation given by the neural ODE?

\begin{figure}[hbt!]
\centering
\includegraphics[width=0.6\textwidth]{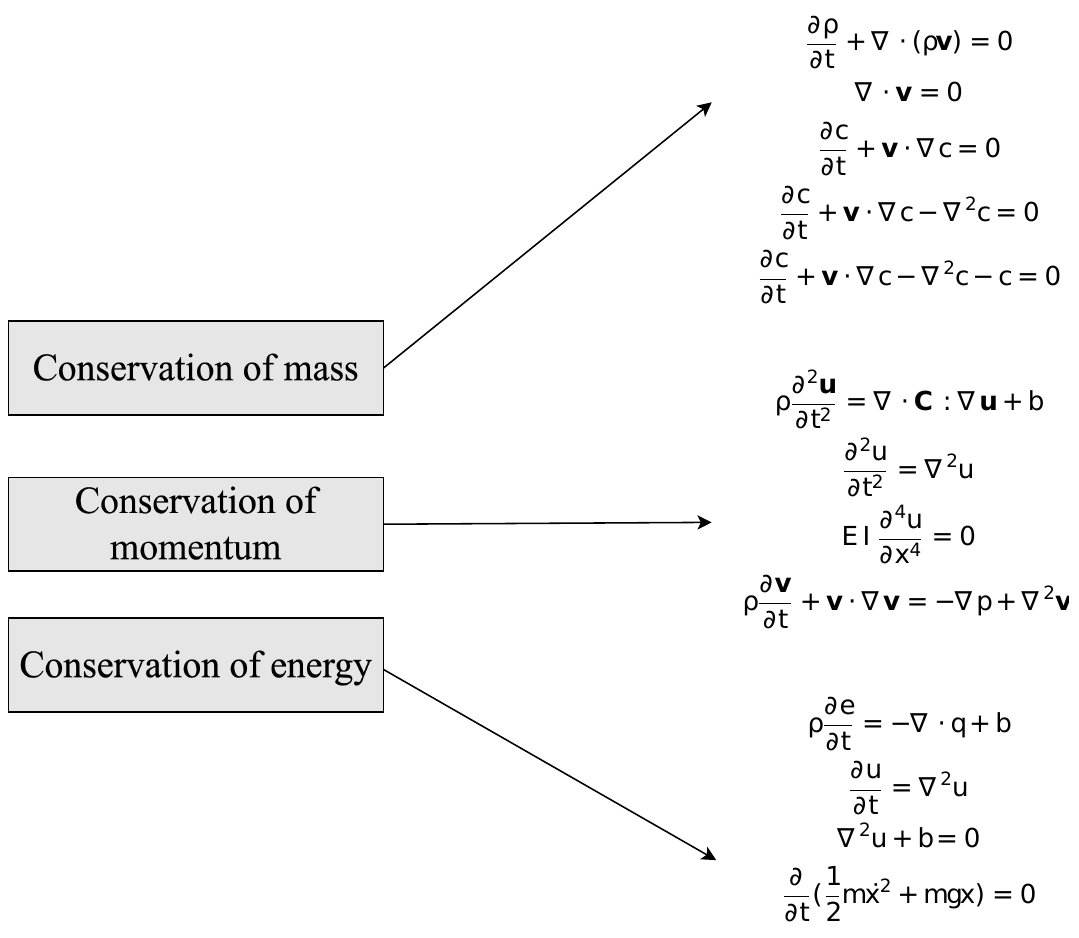}
\caption{Interpreting equations reduces them to fundamental physical principles. The physical principle is analogous to a latent variable that decodes to different equations for different physical systems.}
\label{latent}
\end{figure}

\paragraph{} To answer this question, we remind the reader that we have argued interpretability is important for science because it allows for the unification of apparently disparate phenomena under the banner of a few basic principles. This is Kitcher's idea that good scientific explanations unify \cite{kitcher_explanatory_1981}. In the language of modern ML, and using our definition of interpretation, interpreting an equation is mentally representing it in a maximally compressed latent space, as illustrated in Figure \ref{latent}. If it can be interpreted, Eq. \eqref{SINDy} is not some isolated physical principle of its own, but rather a particular expression of mass conservation in a continuum fluid. How can we generate explanations and interpretations of this type when prior knowledge is lacking? This is a complex question about how scientific theories and models are constructed. The answer philosophers of science typically give is that it is a murky combination of creativity, intuition, luck, and other "irrational" factors \cite{kuhn_structure_2012}. Questions of theory construction aside, what is clear is that---if we aspire to devise ML methods for useful, integrated, novel scientific knowledge---pursuing sparsity is no guarantee of this. Though it is not synonymous with interpretation, might sparsity \textit{assist} interpretation, where interpretation is understood as pulling equations back to the latent space of fundamental physical principles?

\paragraph{}  We claim that the value of sparsity in the absence of prior knowledge is that it leaves the door open for future interpretation. Given that the neural ODE comprises a potentially long composition of affine transformations passed through nonlinear activation functions, there is little hope of recovering this expression in a derivation that begins from physical principles. For whatever reason, history has shown that functions of this sort rarely appear in the governing equations of physics. On the other hand, even in the case when this equation is unfamiliar, it is conceivable that some derivation could terminate in Eq. \ref{SINDy}. If the equation can be derived, the meaning and mechanisms of the various terms become clear, and the equation becomes interpretable. Of course, a sparse equation does not indicate how to reverse engineer its derivation, but sparsity increases the likelihood of some future derivation leading to the discovered equation. With this in mind, the example of Kepler is now appropriate---the sparsity of his planetary laws allowed them to be derived, offering an important verification of Newton's nascent theory of mechanics. Of course, this discussion only applies to expressions that are derived. Thus, we suggest that the value of sparsity is dubious in the setting of empirical relations such as constitutive laws. 

\section{Conclusion}

\begin{figure}[hbt!]
\centering
\includegraphics[width=0.8\textwidth]{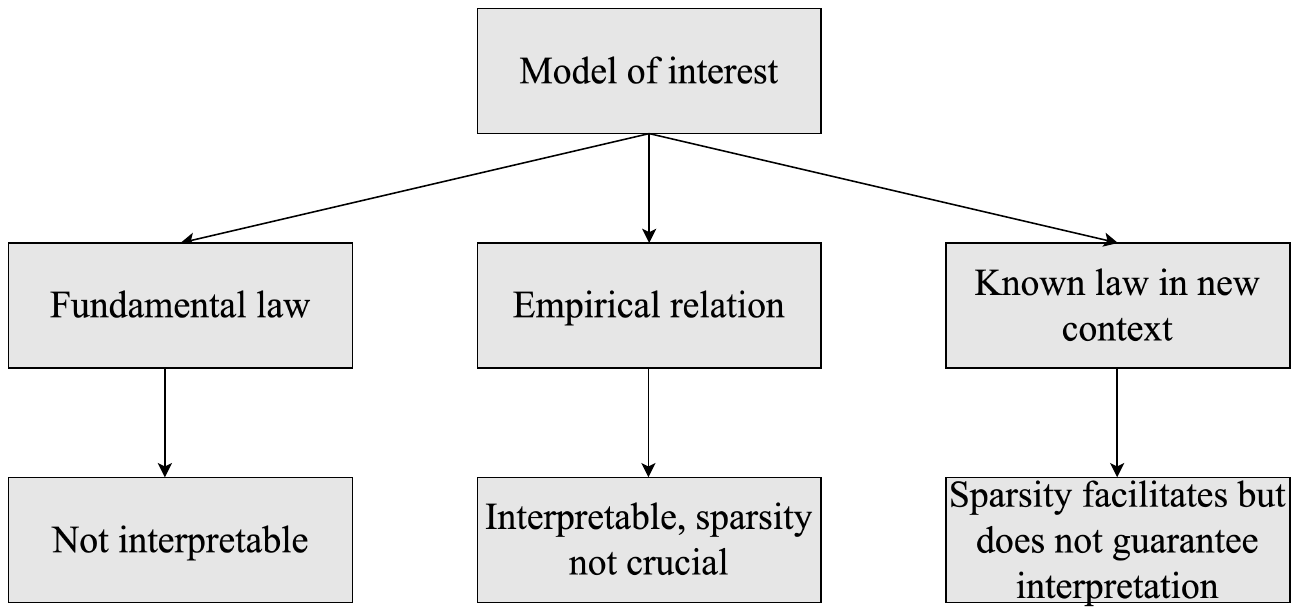}
\caption{Conclusions drawn from investigating the concept of interpretability in SciML. We equate interpretation with the ability to connect terms in an equation to known physical mechanisms. We take physical mechanisms to include both terms in a differential equation with known meaning (advection, diffusion, reaction, etc.) and empirical relations connecting quantities in a modeling framework (constitutive relations). }
\label{conclusions}
\end{figure}

\paragraph{} The purpose of this work was to critically investigate the consensus forming around the definition of interpretability, and its equivalence with sparsity in many works in the SciML literature. To this end, we first reviewed discussions of interpretability in the equation discovery and symbolic regression literature. Though we found some disagreement regarding the definition, our primary finding was that the definition of interpretation as sparsity was most common, but was given implicitly and inadequately justified. In hopes of refining the definition of interpretability for the SciML community, we reviewed some prior work in the broader interpretable ML literature. Here, we found motivations for interpretability that differed significantly from those of research in the physical sciences. In defining interpretability, authors in this literature tended to equate interpretation with the transparency of a model's input-output relation. This definition is not appropriate for science, as scientists look not just for transparent mathematical relationships, but connection to underlying physical principles. In response to vagueness in the SciML literature and the inadequacy of definitions in the interpretable ML literature, we set out to answer two basic questions: What is interpretability in the context of science? And how does interpretability advance scientific knowledge? Despite its popularity, we ruled out sparsity as an adequate definition of interpretability with various counterexamples. We proposed a definition of interpretability that concerns itself with the physical meaning of an equation, as opposed to its mathematical form. We believe this to be more closely aligned with scientists' intuitive understanding of the concept of interpretability. Finally, with the idea that physical principles are a latent space to which disparate physical phenomena decode, we argued that by pulling back to this latent space, interpretability is useful in science as a tool for unification. This argument rescued sparsity from neglect, as sparsity facilitates the derivation of an equation from more fundamental physical principles. However, not all expressions in physics can be derived, so we suggested that sparsity is most useful in the broader task of unification in science for expressions that flow from more fundamental principles, such as conservation laws. Our conclusions are summarized in Figure \ref{conclusions}.

\paragraph{} It is interesting to imagine that the scientific community may opt to integrate neural networks as basic objects of scientific understanding in the future, free of the connection to mechanism we have argued for. However, at this moment, this strategy is not pursued by any research community we are aware of. There is extensive work to make sense of the behavior of networks in terms of more familiar concepts, as Section 3 of our manuscript shows. As such, we take a similar approach, viewing neural networks as something in need of explanation/interpretation. We grant that scientists may forego this approach at some point in the future. Future work will explore in more depth the question of whether interpretations can be extracted from recovered equations when relevant prior knowledge is lacking. Such analysis will allow us to offer an opinion as to the viability of equation discovery and symbolic regression techniques for discovering genuinely novel physical phenomena.

\section*{Declarations}

\subsection*{Acknowledgments}
\paragraph{} This work was funded by the National Defense Science and Engineering Graduate Fellowship (NDSEG) through the Department of Defense (DOD) and the Army Research Office (ARO). A. Doostan was supported by NASA NSTRI and by the US Department of Energy’s Wind Energy Technologies Office.

\subsection*{Competing interest}
\paragraph{} The authors have no relevant financial or non-financial interests to disclose.

\subsection*{Data availability}

\paragraph{} This article is purely conceptual and therefore does not involve the generation, collection, or analysis of data. No datasets or code are associated with this work.



\end{document}